\documentclass{midl}

\jmlrproceedings{MIDL}{Medical Imaging with Deep Learning}
\jmlrpages{}
\jmlryear{2026}

\jmlrworkshop{Short Paper Track}
\jmlrvolume{}
\editors{}

\title[OpenMedQ]{OpenMedQ: Broad Open Pretraining for Medical Vision-Language Models}

\midlauthor{\Name{Ibrahim Gulluk\midljointauthortext{Equal contribution}\nametag{$^{1}$}} \Email{gulluk@stanford.edu}\\
\Name{Max {Van Puyvelde}\midlotherjointauthor\nametag{$^{2,3}$}} \Email{maxvpuyv@stanford.edu}\\
\Name{Olivier Gevaert\nametag{$^{2}$}} \Email{ogevaert@stanford.edu}\\
\addr $^{1}$ Department of Electrical Engineering, Stanford University\\
\addr $^{2}$ Department of Biomedical Data Science, Stanford University School of Medicine\\
\addr $^{3}$ Department of Mathematical Modelling, Statistics and Bioinformatics, Ghent University}

\begin{document}

\maketitle

\begin{abstract}
We present \emph{OpenMedQ}, a medical vision-language model pretrained on the broadest fully-open medical mix to date: 14 datasets totaling ${\sim}3.35$M pretraining samples spanning pathology, radiology, microscopy, and text-only clinical QA. OpenMedQ reaches state-of-the-art BLEU-1 on PathVQA (75.9), beating Med-PaLM~M variants up to 562B parameters (${\sim}80\times$ larger), and matches the best reported VQA-MED BLEU-1 (64.5). Its vision encoder, transferred to 8 unseen medical classification benchmarks under an identical downstream recipe, obtains the highest average macro-F1 (0.757) among BiomedCLIP (0.745), PMC-CLIP (0.745), PubMedCLIP (0.746), and a from-scratch baseline (0.616). We release our \href{https://github.com/gevaertlab/OpenMedQ}{code} 
and an interactive demo is publicly available as a reproducible baseline for the community.
\end{abstract}


\section{Introduction}
Medical foundation models are increasingly capable, yet most published medical VLMs rely on a handful of narrow pretraining sources and withhold either their weights, their data, or both. Contrastive encoders such as BiomedCLIP~\citep{biomedclip}, PMC-CLIP~\citep{pmcclip}, and PubMedCLIP train on single image-caption corpora; generative medical VLMs such as PMC-VQA~\citep{pmcvqa} and LLaVA-Med~\citep{llavamed} demonstrate strong visual question answering (VQA) on a few benchmarks but use comparably narrow pretraining mixes, while BiomedGPT~\citep{biomedgpt} and Med-PaLM~M~\citep{medpalm} scale data and parameters but do not release weights. This leaves practitioners without a fully-open, broadly-pretrained baseline they can actually inspect, reuse, and extend.

We introduce \emph{OpenMedQ}, a LLaVA-style~\citep{llava} VLM (ViT-base~\citep{biomedclip} + LLaMA-7B~\citep{llama,pmcllama}, LoRA~\citep{lora}) trained on the broadest open medical pretraining mix to date (14 datasets, ${\sim}3.35$M samples) with next-token prediction. We will release weights and dataset recipes upon acceptance; a live interactive demo is already available at \url{https://openmedq.streamlit.app/} for qualitative inspection.

\section{Method}
\paragraph{Architecture and pretraining.} The vision encoder $f_{\mathrm{vis}}$ is a ViT-base-patch16-224 initialized from BiomedCLIP~\citep{biomedclip}; a linear projection feeds its image tokens into a LLaMA-7B~\citep{llama} language model initialized from PMC-LLaMA~\citep{pmcllama}. Image and text tokens are concatenated and decoded left-to-right, following LLaVA~\citep{llava}. We fine-tune with LoRA~\citep{lora} of rank $r=8$ using next-token cross-entropy with image and prefix tokens masked. All images are resized to $224{\times}224$; training uses AdamW, batch size 64, learning rate $5{\times}10^{-5}$, for up to 15 epochs on a single NVIDIA A100.

\paragraph{Classification transfer.} To probe the vision features produced by pretraining, we detach $f_{\mathrm{vis}}$ and attach a linear head $W\!\in\!\mathbb{R}^{2d\times m}$; encoder and head are fine-tuned together on each downstream dataset for 100 epochs. We benchmark OpenMedQ's encoder against three strong medical contrastive baselines (BiomedCLIP, PMC-CLIP, PubMedCLIP) and a from-scratch baseline, all under an identical downstream recipe so that any gap is attributable to the pretraining.

\section{Datasets}
\label{sec:data}
\paragraph{Pretraining mix (14 datasets, $\sim$3.35M samples).} Image-text sources (${\sim}2.94$M pairs) span pathology (PathVQA~\citep{pathvqa}), radiology (VQA-RAD~\citep{vqarad}, IU-XRAY~\citep{iuxray}, MIMIC-CXR~\citep{mimiccxr}, ROCO~\citep{roco}, OmniMedVQA~\citep{omnimedvqa}), mixed modalities (Slake~\citep{slake}, PMC-OA~\citep{pmcclip}, PMC-VQA~\citep{pmcvqa}, VQA-MED~\citep{vqamed}), and microscopy ($\mu$-Bench~\citep{ubench}). A further ${\sim}410$K text-only clinical QA samples (MedQA, MedMCQA, PubMedQA) are included to preserve language capability during pretraining.

\paragraph{Classification benchmarks (8 datasets).} We evaluate on CXR8~\citep{cxr8}, MedFMC~\citep{medfmc} (chest, colon, endo subtasks), Breast-Ultrasound~\citep{breastus}, CHAOYANG~\citep{chaoyang}, CBIS-DDSM~\citep{cbisddsm}, and Mendeley-CXray~\citep{mendeley}. These datasets were not seen during pretraining.

\section{Results}
\begin{figure}[t]
\floatconts
  {fig:hero}%
  {\caption{\textbf{(a)} Macro-F1 across 8 unseen medical classification benchmarks: all bars share an identical downstream recipe and differ only in the pretrained vision encoder. OpenMedQ attains the highest \emph{Mean} (0.757). \textbf{(b)} OpenMedQ's pretraining mix: 14 fully-open datasets (${\sim}3.35$M pairs), colored by modality group.}\label{fig:hero}}%
  {\includegraphics[width=\linewidth]{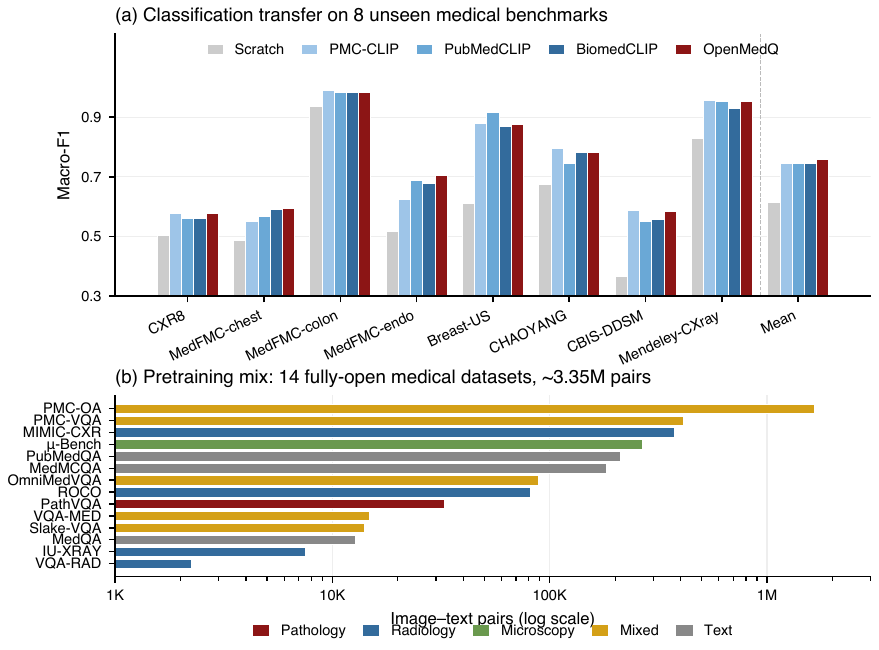}}
\end{figure}

\paragraph{Classification transfer.} \figureref{fig:hero}(a) is our headline result. OpenMedQ achieves the highest mean macro-F1 (0.757) across the eight benchmarks, ahead of PubMedCLIP (0.746), PMC-CLIP and BiomedCLIP (0.745), and the from-scratch baseline (0.616). OpenMedQ wins outright on MedFMC-chest and MedFMC-endo, ties PMC-CLIP on CXR8, and trails the best encoder by at most 0.02 on four more; the only meaningful gap is Breast-Ultrasound (0.876 vs.\ 0.915). Since the downstream recipe is fixed, this delta reflects what OpenMedQ's pretraining added to the BiomedCLIP initialization.

\paragraph{Open-ended VQA.} \begin{sloppypar}On PathVQA, OpenMedQ reaches 75.9 BLEU-1, beating prefix tuning~\citep{vansonsbeek} (70.3) and all three Med-PaLM~M variants up to 562B~\citep{medpalm} (72.27) despite using only 7B parameters. On VQA-MED, OpenMedQ reaches 64.5, just above the 2019 challenge best (64.4).\end{sloppypar}

\section{Discussion}
\emph{Breadth} of open pretraining data is a competitive lever for medical VLMs: at 7B parameters, OpenMedQ sets a new state of the art on PathVQA against Med-PaLM~M up to 562B, and its vision encoder beats three strong contrastive medical encoders on average classification transfer. Data diversity is a reproducible lever; proprietary scale is not. The lever has its limits: Med-PaLM~M's larger variants still lead on VQA-RAD and Slake, BLEU-1 captures only surface agreement, and narrow-modality encoders can edge us out on Breast-Ultrasound. The demo is available at \url{https://openmedq.streamlit.app/}.

% Acknowledgments--Will not appear in anonymized version
%\midlacknowledgments{}

\bibliography{refs}

@inproceedings{llava,
  title={Visual instruction tuning},
  author={Liu, Haotian and Li, Chunyuan and Wu, Qingyang and Lee, Yong Jae},
  booktitle={Advances in Neural Information Processing Systems},
  volume={36},
  year={2024}
}

@article{llama,
  title={{LLaMA}: Open and efficient foundation language models},
  author={Touvron, Hugo and Lavril, Thibaut and Izacard, Gautier and Martinet, Xavier and Lachaux, Marie-Anne and Lacroix, Timoth{\'e}e and Rozi{\`e}re, Baptiste and Goyal, Naman and Hambro, Eric and Azhar, Faisal and others},
  journal={arXiv preprint arXiv:2302.13971},
  year={2023}
}

@article{pmcllama,
  title={{PMC-LLaMA}: Toward building open-source language models for medicine},
  author={Wu, Chaoyi and Lin, Weixiong and Zhang, Xiaoman and Zhang, Ya and Xie, Weidi and Wang, Yanfeng},
  journal={Journal of the American Medical Informatics Association},
  volume={31},
  number={9},
  pages={1833--1843},
  year={2024}
}

@article{lora,
  title={{LoRA}: Low-rank adaptation of large language models},
  author={Hu, Edward J and Shen, Yelong and Wallis, Phillip and Allen-Zhu, Zeyuan and Li, Yuanzhi and Wang, Shean and Wang, Lu and Chen, Weizhu},
  journal={arXiv preprint arXiv:2106.09685},
  year={2021}
}

@article{biomedclip,
  title={{BiomedCLIP}: A multimodal biomedical foundation model pretrained from fifteen million scientific image-text pairs},
  author={Zhang, Sheng and Xu, Yanbo and Usuyama, Naoto and Xu, Hanwen and Bagga, Jaspreet and Tinn, Robert and Preston, Sam and Rao, Rajesh and Wei, Mu and Valluri, Naveen and others},
  journal={arXiv preprint arXiv:2303.00915},
  year={2023}
}

@inproceedings{pmcclip,
  title={{PMC-CLIP}: Contrastive language-image pre-training using biomedical documents},
  author={Lin, Weixiong and Zhao, Ziheng and Zhang, Xiaoman and Wu, Chaoyi and Zhang, Ya and Wang, Yanfeng and Xie, Weidi},
  booktitle={International Conference on Medical Image Computing and Computer-Assisted Intervention},
  pages={525--536},
  year={2023},
  organization={Springer}
}

@article{pmcvqa,
  title={{PMC-VQA}: Visual instruction tuning for medical visual question answering},
  author={Zhang, Xiaoman and Wu, Chaoyi and Zhao, Ziheng and Lin, Weixiong and Zhang, Ya and Wang, Yanfeng and Xie, Weidi},
  journal={arXiv preprint arXiv:2305.10415},
  year={2023}
}

@article{llavamed,
  title={{LLaVA-Med}: Training a large language-and-vision assistant for biomedicine in one day},
  author={Li, Chunyuan and Wong, Cliff and Zhang, Sheng and Usuyama, Naoto and Liu, Haotian and Yang, Jianwei and Naumann, Tristan and Poon, Hoifung and Gao, Jianfeng},
  journal={Advances in Neural Information Processing Systems},
  volume={36},
  year={2024}
}

@article{biomedgpt,
  title={{BiomedGPT}: A unified and generalist biomedical generative pre-trained transformer for vision, language, and multimodal tasks},
  author={Zhang, Kai and Yu, Jun and Yan, Zhiling and Liu, Yixin and Adhikarla, Eashan and Fu, Sunyang and Chen, Xun and Chen, Chen and Zhou, Yuyin and Li, Xiang and others},
  journal={arXiv preprint arXiv:2305.17100},
  year={2023}
}

@article{medpalm,
  title={Towards generalist biomedical {AI}},
  author={Tu, Tao and Azizi, Shekoofeh and Driess, Danny and Schaekermann, Mike and Amin, Mohamed and Chang, Pi-Chuan and Carroll, Andrew and Lau, Charles and Tanno, Ryutaro and Ktena, Ira and others},
  journal={NEJM AI},
  volume={1},
  number={3},
  pages={AIoa2300138},
  year={2024}
}

@article{pathvqa,
  title={{PathVQA}: 30000+ questions for medical visual question answering},
  author={He, Xuehai and Zhang, Yichen and Mou, Luntian and Xing, Eric and Xie, Pengtao},
  journal={arXiv preprint arXiv:2003.10286},
  year={2020}
}

@article{vqarad,
  title={A dataset of clinically generated visual questions and answers about radiology images},
  author={Lau, Jason J and Gayen, Soumya and Ben Abacha, Asma and Demner-Fushman, Dina},
  journal={Scientific Data},
  volume={5},
  number={1},
  pages={1--10},
  year={2018}
}

@inproceedings{slake,
  title={{SLAKE}: A semantically-labeled knowledge-enhanced dataset for medical visual question answering},
  author={Liu, Bo and Zhan, Li-Ming and Xu, Li and Ma, Lin and Yang, Yan and Wu, Xiao-Ming},
  booktitle={IEEE International Symposium on Biomedical Imaging (ISBI)},
  pages={1650--1654},
  year={2021}
}

@inproceedings{vqamed,
  title={{VQA-Med}: Overview of the medical visual question answering task at {ImageCLEF} 2019},
  author={Ben Abacha, Asma and Hasan, Sadid A and Datla, Vivek V and Demner-Fushman, Dina and M{\"u}ller, Henning},
  booktitle={CLEF Working Notes},
  year={2019}
}

@article{iuxray,
  title={Preparing a collection of radiology examinations for distribution and retrieval},
  author={Demner-Fushman, Dina and Kohli, Marc D and Rosenman, Marc B and Shooshan, Sonya E and Rodriguez, Laritza and Antani, Sameer and Thoma, George R and McDonald, Clement J},
  journal={Journal of the American Medical Informatics Association},
  volume={23},
  number={2},
  pages={304--310},
  year={2016}
}

@article{mimiccxr,
  title={{MIMIC-CXR}, a de-identified publicly available database of chest radiographs with free-text reports},
  author={Johnson, Alistair EW and Pollard, Tom J and Berkowitz, Seth J and Greenbaum, Nathaniel R and Lungren, Matthew P and Deng, Chih-ying and Mark, Roger G and Horng, Steven},
  journal={Scientific Data},
  volume={6},
  number={1},
  pages={317},
  year={2019}
}

@inproceedings{roco,
  title={Radiology Objects in COntext ({ROCO}): A multimodal image dataset},
  author={Pelka, Obioma and Koitka, Sven and R{\"u}ckert, Johannes and Nensa, Felix and Friedrich, Christoph M},
  booktitle={Intravascular Imaging and Computer Assisted Stenting and Large-Scale Annotation of Biomedical Data and Expert Label Synthesis},
  pages={180--189},
  year={2018}
}

@inproceedings{omnimedvqa,
  title={{OmniMedVQA}: A new large-scale comprehensive evaluation benchmark for medical {LVLM}},
  author={Hu, Yutao and Li, Tianbin and Lu, Quanfeng and Shao, Wenqi and He, Junjun and Qiao, Yu and Luo, Ping},
  booktitle={Proceedings of the IEEE/CVF Conference on Computer Vision and Pattern Recognition},
  pages={22170--22183},
  year={2024}
}

@article{ubench,
  title={{$\mu$-Bench}: A vision-language benchmark for microscopy understanding},
  author={Lozano, Alejandro and Nirschl, Jeffrey and Burgess, James and Gupte, Sanket Rajan and Zhang, Yuhui and Unell, Alyssa and Yeung-Levy, Serena},
  journal={arXiv preprint arXiv:2407.01791},
  year={2024}
}

@inproceedings{cxr8,
  title={{ChestX-ray8}: Hospital-scale chest x-ray database and benchmarks on weakly-supervised classification and localization of common thorax diseases},
  author={Wang, Xiaosong and Peng, Yifan and Lu, Le and Lu, Zhiyong and Bagheri, Mohammadhadi and Summers, Ronald M},
  booktitle={Proceedings of the IEEE Conference on Computer Vision and Pattern Recognition},
  pages={2097--2106},
  year={2017}
}

@misc{medfmc,
  title={{MedFM} 2023 Grand Challenge: {MedFM} 2023 datasets},
  howpublished={\url{https://medfm2023.grand-challenge.org/datasets/}},
  note={Accessed 2024-07-30},
  year={2023}
}

@article{breastus,
  title={Dataset of breast ultrasound images},
  author={Al-Dhabyani, Walid and Gomaa, Mohammed and Khaled, Hussien and Fahmy, Aly},
  journal={Data in Brief},
  volume={28},
  pages={104863},
  year={2020}
}

\end{document}